\documentclass[10pt,twocolumn,letterpaper]{article}

\usepackage[utf8]{inputenc} 
\usepackage[T1]{fontenc}    
\usepackage{booktabs}       
\usepackage{amsfonts}       
\usepackage{nicefrac}       
\usepackage{microtype}      
\usepackage{amsmath}
\usepackage{multicol}
\usepackage{multirow}
\usepackage{array}
\usepackage{amsmath}
\usepackage{algorithm}
\usepackage{algpseudocode}
\usepackage{graphicx}
\usepackage{amssymb}
\usepackage{graphics}
\usepackage{caption}
\usepackage{authblk}

\usepackage{cvpr}              

\usepackage[dvipsnames]{xcolor}

\definecolor{cvprblue}{rgb}{0.21,0.49,0.74}
\usepackage[pagebackref,breaklinks,colorlinks,citecolor=cvprblue]{hyperref}

\def\paperID{2340} 
\def\confName{CVPR}
\def\confYear{2024}

\algnewcommand{\IfThenElse}[3]{
  \State \algorithmicif\ #1\ \algorithmicthen\ #2\ \algorithmicelse\ #3}

\title{Progressive Energy-Based Cooperative Learning for  Multi-Domain Image-to-Image Translation}

\author[1]{Weinan Song}
\author[1]{Yaxuan Zhu}
\author[2,1]{Lei He}
\author[1]{Ying Nian Wu}
\author[3]{Jianwen Xie}
\affil[1]{University of California, Los Angeles}
\affil[2]{Eastern Institute for Advanced Study, Ningbo, China}
\affil[3]{Akool Research}

\begin{document}
\maketitle

\begin{abstract}
This paper studies a novel energy-based cooperative learning framework for multi-domain image-to-image translation. The framework consists of four components: descriptor, translator, style encoder, and  style generator. The descriptor is a multi-head energy-based model that represents a multi-domain image distribution. Given an image from a source domain, the translator generates an output image in the target domain given a style code, which can be inferred by the style encoder from a reference image or generated by the style generator from a random noise. To train our framework, we propose a likelihood-based multi-domain cooperative learning algorithm to jointly train the multi-domain descriptor and the diversified image generator (including translator, style encoder, and style generator) via multi-domain MCMC teaching, in which the descriptor guides the diversified image generator to shift its probability density toward the data distribution, while the diversified image generator uses its randomly translated images to initialize the descriptor's Langevin dynamics process for efficient sampling. We also bring in two regularization strategies for both the descriptor and the translator to significantly improve the cooperative learning. To further enhance the efficiency and scalability, we propose a progressive cooperative learning strategy to train our framework. Strong empirical results are shown to verify the effectiveness of our energy-based image translation~framework.
\end{abstract}

\section{Introduction}
The task of image-to-image translation primarily involves the learning of  mappings between different visual domains. This learning framework carries immense application value in the field of generative artificial intelligence, facilitating the development of various creative products for artificial intelligence-generated contents (AIGC). In this context, a ``domain'' refers to a collection of images belonging to a visually distinctive category such as the gender of a person and animal species. Within each domain, every image exhibits a unique appearance, encompassing image-specific elements such as hairstyle and makeup, commonly referred to as its ``style''. An ideal image-to-image translation framework should possess the ability to handle multiple domains, efficiently process high-resolution images, and provide diverse synthesis (i.e., one-to-many mapping) when translating to each target domain. By leveraging the representation power of an energy-based model and the sampling efficiency of a latent variable model, the Generative Cooperative Network \citep{coopnet}, also known as CoopNets, and its variants \citep{XieZL21, XieZLL22}, have achieved impressive results in numerous computer vision tasks, such as image generation \citep{coopnet, XieZL21, XieZLL22}, visual salient object detection\citep{ZhangXZB22}, supervised image-to-image translation \citep{XieZFZW22}, and unsupervised image-to-image translation \citep{coop_cycle}. However, while the cross-domain translation framework, CycleCoopNets \citep{coop_cycle}, has demonstrated success in unpaired image-to-image translation, it is only capable of learning the relation between two different domains at a time. Such an approach has a limited scalability to deal with multiple domains, as a separate model must be trained for each pair of domains. Besides, cooperative learning still faces challenges when it comes to translating high-resolution images. This is because the translation process involves sampling from the energy-based model via Langevin dynamics, which can be difficult to apply to high-resolution image spaces. To tackle the aforementioned challenges in the current cooperative learning (or more generally, energy-based learning) for multi-domain unsupervised image-to-image translation, this paper proposes a novel cooperative learning framework, PMD-CoopNets, to ensure \textbf{scalability}, \textbf{flexibility}, \textbf{stability} and \textbf{efficiency} for applying energy-based framework to image-to-image translation. 

To be specific, the PMD-CoopNets consists of four components: descriptor, translator, style generator and style encoder. (1) The descriptor is a multi-head energy-based model that represents a multi-domain image distribution, where each head of the energy function corresponds to one image domain. (2) The style generator is a multi-head latent variable model responsible for generating domain-specific style codes.  It achieves this by transforming a Gaussian latent code into style codes. Each head in the style generator corresponds to one specific domain. (3) The style encoder extracts domain-specific style codes from an input image using a multi-head encoder. Each head of the encoder corresponds to a specific domain. (4) The translator is a style-controlled mapping, which takes an image and a style code as input, and then transforms the image into a translated image that reflects the desired style indicated by the style code. The style code can be obtained either from the style generator or the style extractor. The style generator, style encoder, and style-controlled translator can constitute a diversified image generator. 

As to the learning, the multi-domain descriptor and the diversified image generator engage in a cooperative game, where the multi-domain descriptor guides the diversified image generator in aligning its mapping towards the target domains using MCMC teaching, while the image generator assists in expediting the descriptor's MCMC teaching process by providing a good initialization. Specifically, to enforce a meaningful latent space of style codes, we train the style-controlled translator and style encoder by reconstructing style codes that are randomly generated from the style generator. To enforce translated image to preserve the domain-invariant property of the input reference image, we train the translator with a cycle consistency loss. To enforce the one-to-many translation output, we regularize the translator via a diversity sensitive loss, such that, given an identical reference image, different style codes can lead to sufficiently diversified translated outputs.

Additionally, we propose to improve the cooperative learning algorithm by incorporating some loss terms to regularize the behaviors of the components in our framework. Firstly, we put an $l_2$ regularization on the output of the energy function of the descriptor to limit the magnitude of the energy values. To accelerate and stabilize the teaching process provided by the descriptor's MCMC, we propose to use the energy function to regularize the output of the translator. These regularization techniques significantly improve the performance of the cooperative learning.         

To enhance efficiency, stability, and scalability, we present a progressive cooperative learning algorithm for our model. Our approach involves gradual expansion of all four components, initially operating on simpler low-resolution images. As the cooperative training proceeds, new layers are added to each component, enabling the model to handle more challenging high-resolution images. This progressive growth strategy significantly accelerates and stabilizes both training and sampling processes at higher resolutions. Moreover, it offers the flexibility and convenience to scale up the resolution of any pre-trained PMD-CoopNets.

We demonstrate the effectiveness of our proposed multi-domain translation model on the CelebA-HQ \citep{progressive_gan} and AFHQ \citep{stargan2} dataset. The translated examples exhibit high fidelity and are comparable to GAN-based multi-domain translation models in the task of image-to-image translation. Furthermore, our progressive learning strategy improves the efficiency and stability of the original training process, particularly when it comes to translating high-resolution images. Our contributions are listed below:
\begin{itemize}
    \item We propose a novel energy-based cooperative learning framework for multi-domain image-to-image translation. We build a single multi-head energy-based model to represent probability distributions of multiple domains, and train it with a translator, a style encoder, and a style generator using a cooperative manner.   
    \item We present a novel progressive learning algorithm to optimize the training efficiency of our framework. Our approach adopts a progressive growth strategy, advancing all components from low resolution to high resolution. It yields a significant reduction in the total number of MCMC steps required for training and sampling from the high-resolution model.
    \item We propose regularization strategies to stabilize the cooperative learning, which include an energy-based regularization loss for the translator and a $l_2$ regularization loss for limiting the magnitude of the energy values of the descriptor. Significant performance gain are obtained from these regularization.  
    \item We demonstrate strong empirical results on CelebA-HQ and AFHQ datasets to verify the proposed energy-based framework. Our method obtains state-of-the-art performance among existing energy-based image translation models.
\end{itemize}
\section{Related Work}

\paragraph{Energy-based Learning}
Training energy-based models (EBMs) \cite{zhu1998filters, lecun2006tutorial, lncs_Hinton12} involves maximizing the likelihood of the observed data by adjusting the model's energy function parameters, which typically requires Markov chain Monte Carlo (MCMC) sampling to evaluate the intractable gradient \cite{XieLZW16, nijkamp2019learning, du2019implicit}. Contrastive divergence (CD) \cite{hinton2002training, du2020improved} is an efficient approximation algorithm for training energy-based models by initializing the MCMC chains with observed data. \cite{nijkamp2019learning} uses a noise-initialized non-convergent short-run MCMC to train an EBM, and obtains a valid flow-like generator trained with moment matching estimation. \cite{GaoSPWK21} defines a sequence of conditional EBMs and forms a denoising diffusion process. 
To avoid MCMC, \cite{gao2020flow} brings in normalizing flow and trains an EBM by flow contrastive estimation. Learning an amortized sampler \cite{kim2016deep, coopnet, HanNFHZW19, kumar2019maximum, xiao2020vaebm,  GrathwohlKH0SD21} for EBMs is also an alternative strategy. Our method has a single multi-head EBM to represent multi-domain data distribution, and the image-to-image translator serves as a multi-domain amortized sampler for the EBMs.        

\paragraph{Cooperative Learning}
Cooperative learning for energy-based models with MCMC teaching is first proposed in \cite{coopnet}, where the authors utilize an energy-based model as the descriptor and a latent variable model as the generator to speed up the learning of each other by maximum likelihood algorithms. During each training iteration, the descriptor generates samples by ﬁnite-step MCMC sampling with initialization by the generation from the generator for maximum likelihood estimation. Simultaneously, the sampling results from descriptor are used to directly supervise the generator, which is called MCMC teaching. Further research in \cite{coopinit} shows that this cooperative learning method could also provide a good start point for adversarial models with small computation overhead. Additionally, the model could also be extended for image-to-image translation \cite{coop_cycle} with two pairs of descriptor and generator or used in saliency prediction \cite{salcoopnet} by introducing a conditional latent variable model.

\paragraph{Progressive Learning}
The proposed idea of progressive cooperative learning is closely connected to the research conducted by \cite{zhao2020learning}, which involves the incremental growth of a single EBM. The multi-grid EBM framework \cite{GaoLZZW18}, trains a series of EBMs simultaneously at various resolutions. The sampling process is conducted sequentially, starting from low-resolution and gradually progressing to higher resolutions, leveraging the lower resolution as a foundation for subsequent higher-resolution sampling. In contrast, our method, which combines the growth of an EBM with three mapping networks, introduces a more challenging and complex progressive learning strategy. It is important to note that while there are several progressive learning frameworks based on Generative Adversarial Networks (GANs), our approach falls within the domain of energy-based learning. We need to carefully consider MCMC sampling when progressively expanding the energy function, as it plays a crucial role in both bottom-up energy mapping and top-down image generation.


\section{Proposed Framework}
\begin{figure*}[t]
    \centering
    \includegraphics[width=0.85\textwidth]{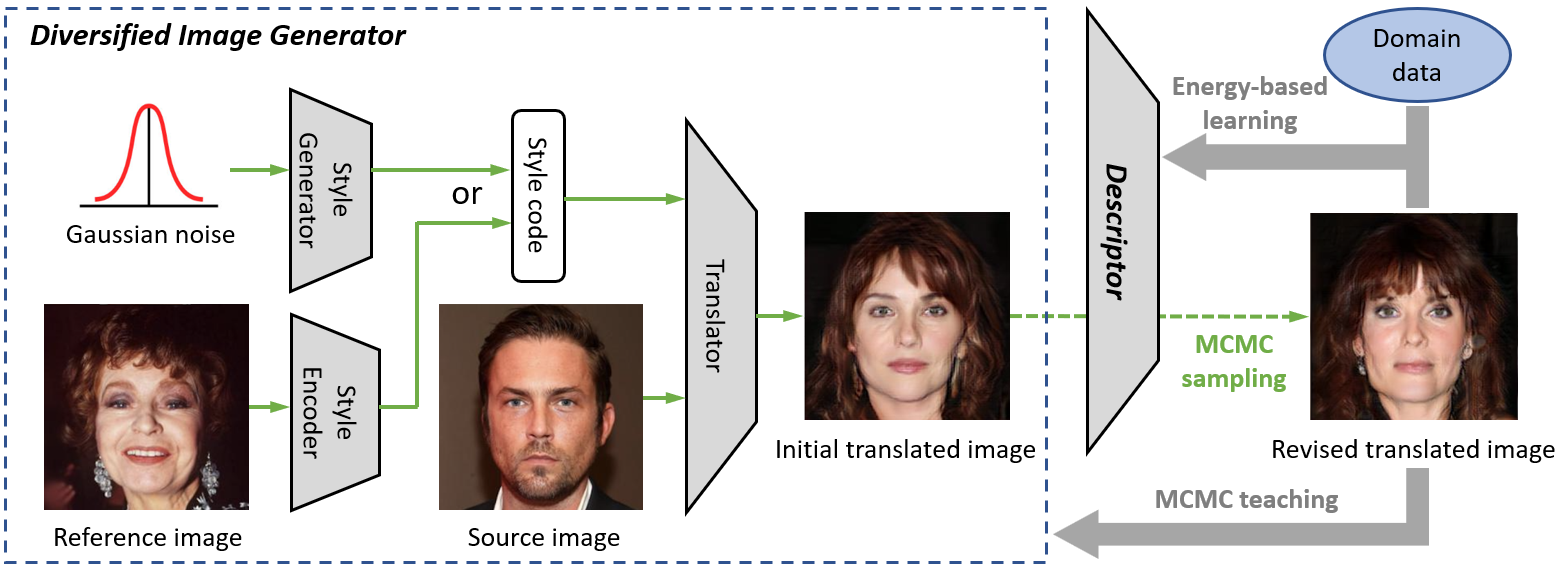}
    \label{fig:psc_overview}
    \caption{Diagram of energy-based cooperative learning for multi-domain image-to-image translation. The
framework consists of a style generator, a style encoder, a translator and a descriptor. The first three components (i.e., style generator, style encoder, and translator) form a diversified image generator. Given a input source image, the translator can transform it into a target domain, which is specified by a style code. The style code can be obtained by sampling from the domain-specific style generator or extracted from a reference image by the style encoder. The descriptor is a multi-domain image distribution, which plays the role of guiding the translation such that the translated images can match the observed
images in the target domain in terms of statistical property. All components are trained simultaneously in a cooperative learning scheme. The descriptor learns from the multi-domain training images by maximizing the data likelihood, while utilizing MCMC teaching to guide the training of the diversified image generator, which consists of a translator,
a style encoder, and a style generator.
    }
\end{figure*}

Suppose we have unpaired images from multiple domains A, B, C, 
$\cdots$ with some shared high-level features, such as expressions in face images, our target is to learn a conditional generative model that maps an image into a target domain, which could be same as the source domain, with specific features. To achieve this, we propose a generative model that consists of four components, i.e., descriptor, style encoder, style generator,  and translator. The latter three can form a diversified translator, which is trained with the descriptor in a cooperative learning manner. Let $x$ be an observed image and $y$ be its domain label. We also use $y'$ to denote the label of target domain.  

\subsection{Multi-Domain Descriptor}
The multi-domain descriptor is a multi-head energy-based model that specifies the probability distribution of each domain by
\begin{equation}
p_y(x;\theta)\propto\exp [D_{y}(x;\theta)], 
\end{equation}
where $\theta$ are parameters of the multi-head energy function $D$. For notation simplicity, we use $D_y(\cdot)$ to denote the negative energy for domain $y$.
The descriptor are learned by multi-domain maximum likelihood estimation, which is equivalent to minimizing the Kullback-Leibler (KL) divergence between the data distribution $p_\text{data}(x,y)$ and the model $p_y(x;\theta)$.  The gradient of the objective for learning the descriptor is given by
\label{eq:ebm_grad}
\begin{align}
\begin{split}
&\nabla_{\theta} \mathcal{L}_{\text{ebm}}(\theta) \\
&= -\mathbb{E}_{p_\text{data}(x,y)}\{\nabla_{\theta} D_{y}(x; \theta) -\mathbb{E}_{p_{y}(x';\theta)}\left[\nabla_{\theta} D_{y}\left(x'; \theta \right)\right]\},
\end{split}
\end{align}
where $\mathbb{E}_{p_y(x';\theta)}$ denotes the expectation with respect to the EBM and we use $x'$ in order to distinguish the random variable $x$ in $\mathbb{E}_{p_{\text{data}}(x, y)}$ in the same equation. Suppose we observe a batch of training examples $\{(x_i,y_i)\}_i^n$, which is assumed to be from $p_{\text{data}}(x,y)$. The gradient in Eq.(\ref{eq:ebm_grad}) can be approximated~by
\begin{equation}
\label{eq:ebm_grad_appro}\nabla_{\theta} \mathcal{L}_{\text{ebm}}(\theta)\approx \nabla_{\theta} \left[\frac{1}{n} \sum_{i=1}^n  D_{y_i}(x_i; \theta)-\frac{1}{n} \sum_{i=1}^{n}  D_{y_i}(\tilde{x}_i; \theta)\right],
\end{equation}
where for each observed domain $y_i$, we use Langevin dynamics to obtain the corresponding synthesized example $\tilde{x}_i$ as a sample from $p_{y_i}(x;\theta)$. With a specified step size $\delta$, Langevin dynamics is performed by iterating the follow~step:
\begin{equation}
\label{eq: langevin_sampling}
\tilde{x}_{\tau+1}=\tilde{x}_\tau +\delta \nabla_{x} D_{y}\left(\tilde{x}_\tau;\theta\right)+ \sqrt{2\delta} U_\tau, U_\tau \sim \mathcal{N}(0, I),
\end{equation}
where $\tau$ indexes time step and $\tilde{x}_{\tau=0}$ is initialized by the output of a style-controlled image-to-image translator, which is presented in Section \ref{sec:translator}. A good initialization improves the efficiency of Langevin dynamics. To stabilize the EBM training, we also add an $l_2$ regularization on the energy outputs of both training examples and synthesized examples, which is
\begin{equation}
\label{eq: l2_regularization}
\mathcal{L}_{\text{energy}}(\theta)=\frac{1}{n}\sum_{i=1}^n \Vert D_{y_i}(x_i;\theta)\Vert^2 + \frac{1}{n} \sum_{i=1}^{n}\Vert D_{y_i}(\tilde{x}_i;\theta)\Vert^2.
\end{equation}

\begin{figure*}[t]
    \centering
    \includegraphics[width=0.9\textwidth]{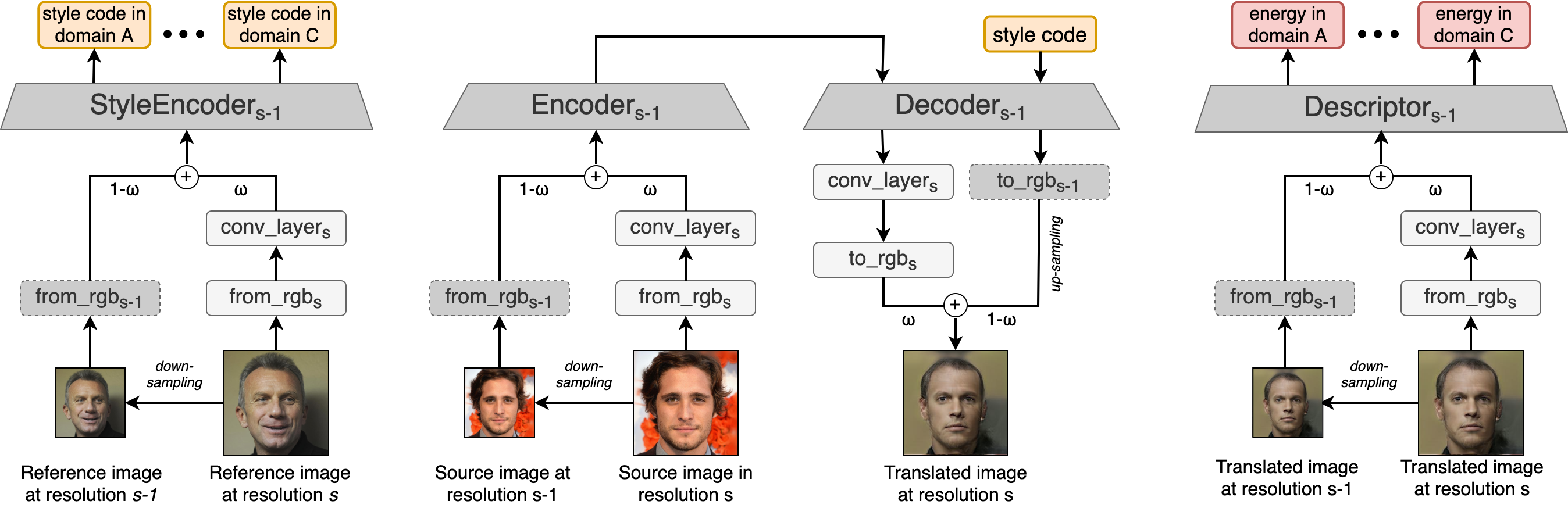}    
    \caption{An illustration of the progressive strategy for the style encoder $E$, translator $T$, and descriptor $D$. Boxes in dark grey represent well-trained modules at resolution level $s-1$, while blocks in light gray represent the newly added parameters at the current resolution level $s$. The expansion of the model involves removing some incompatible parameters (depicted as dark grey boxes with dashed boundaries) and adding new parameters (depicted as light grey boxes). The output of the module that needs to be removed and the output of the module that needs to be added are fused using a transition factor $\omega$, This factor starts from 0 and gradually increases to 1, controlling the percentage of contribution from the old and new modules. Left: style encoder. Middle: style-controlled image-to-image translator. Right: descriptor.
    }\label{fig:psc_progressive}
\end{figure*}

\subsection{Diversified Image Generator}

\paragraph{Multi-Domain Style Generator} 
Given a latent variables $z$ and a domain label $y$, the multi-domain style generator can produce a domain-specific style code $c$ by
\begin{equation}
\label{eq: style_generator}
c_y=G_y(z; \beta)+ \epsilon, \hspace{2mm} \epsilon \sim \mathcal{N}(0, I), z \sim \mathcal{N}(0, I),
\end{equation}
where $\epsilon$ is an observation residual and $z$ follows a Gaussian prior distribution. $G$ is a multilayer perceptron (MLP) with multiple output branches to produce style codes for multiple domains. The distribution of style code $c$ conditioned on a domain $y$ is given by $p_{y}(c;\beta)=\int p_{y}(c|z; \beta)p(z) dz$, which is more informative than the prior distribution $p(z)$ to capture the underlying style space. The domain-specific style code $c_y$ is directly used in the translator, which is presented in Section \ref{sec:translator}, for specifying the style and the target domain of the translated image. 

\paragraph{Style Encoder}
The style encoder $E$ is a multi-head bottom-up network that takes as input an image $x$ and its corresponding domain label $y$ and then outputs a domain-specific style code $c=E_y(x;\phi)$, where $\phi$ are parameters and $E_y(\cdot)$ denotes the output of $E$ that corresponds to domain $y$.

\paragraph{Style-Controlled Image-to-Image Translator}\label{sec:translator}
To achieve a one-to-many translation between domains, we build a style-controlled image-to-image translator. It is a conditioned encoder-decoder $T$ that takes as input a source reference image $x$ and a domain-specific style code $c_y$ and outputs a translated image in target domain $y$, which is given by 
\begin{equation}
\label{eq: translator}
x_y=T(x, c_y; \alpha)+ \epsilon, \hspace{2mm} \epsilon \sim \mathcal{N}(0, I), c_y \sim p_y(c;\beta),
\end{equation}
where $\alpha$ is the parameters of the neural network $T$. The randomness in the translated image, when given a reference image and the target domain, arises from the stochastic nature of the style codes, which follows a distribution defined by the style generator $p_y(c;\beta)$. The translator $T$ and the style generator $G$ forms a diversified translator. They are trained by the MCMC teaching loss \cite{coopnet}, which is
\begin{equation}
\label{eq:MCMC_teaching}
    \mathcal{L}_{\text{teach}}(\alpha, \beta) = \mathbb{E}_{z, y, x} [\Vert \tilde{x}_{z,y,x} - T(x, G_y(z;\beta); \alpha) \Vert^2],
\end{equation}
where $\tilde{x}_{z,y,x}$ denotes the Langevin synthesis from the descriptor, which is initialized by the output of $T(x, G_y(z;\beta); \alpha)$. 
That is, we set $\tilde{x}_{z,x,y,\tau=0} \leftarrow T(x, G_y(z;\beta)$ for Langevin dynamics in Eq.(\ref{eq: langevin_sampling}) to revolve $\tilde{x}_{z,y,x}$. Let $M_{\theta}q_{\alpha,\beta}(x)$ be the marginal distribution obtained by running Markov transition $M_{\theta}$ from $q(x;\alpha,\beta)$. At learning step $t+1$, the gradient of the MCMC teaching loss in Eq.(\ref{eq:MCMC_teaching}) is the gradient of $\text{KL}(M_{\theta^{(t)}}q_{\alpha^{(t)},\beta^{(t)}}||q_{\alpha,\beta})$, where $q_{\alpha,\beta}$ seeks to be the stationary distribution of $M_{\theta}$, i.e., minimizing $\text{KL}(p_{\theta}||q_{\alpha, \beta})$. The effects of the MCMC teaching loss include: (i) $q$ can chase $p$ toward $p_{\text{data}}$ for MLE; (ii) $q$ can serve as a good MCMC initializer for $p$ for efficient MCMC sampling. 
To ensure diverse  translator outputs, we regularize $T$ by minimizing the negative diversity sensitive 
\begin{align}
\begin{split}
\label{eq:diversity_loss}
\mathcal{L}_{\text{diverse}}(\alpha) = -\mathbb{E}_{z_1, z_2, y, x} [\Vert &T(x,G_y(z_1;\beta);\alpha) \\
&- T(x, G_y(z_2;\beta); \alpha) \Vert_1].
\end{split}
\end{align}
Since the translator is learned from unpaired data domains, to ensure the translated image $T(x,c;\alpha)$ to preserve the domain-invariant features of the source image $x$, we adopt the cycle consistency loss:
\begin{equation}
\label{eq:cycle_loss}
\mathcal{L}_{\text{cycle}}(\alpha) = \mathbb{E}_{z, y, x, y'} [\Vert x - x_{cycle} \Vert_1],
\end{equation}
where $x_{cycle} = T( T(x,G_{y'}(z;\beta);\alpha), E_y(x;\phi); \alpha)$. To ensure any style code that is applied to the translated image can be retrieved back from the translated image by the style encoder, we also have a style code reconstruction loss
\begin{align}
\begin{split}
\label{eq:style_code_reconstruction}
&\mathcal{L}_{\text{style}}(\alpha,\phi) \\
&=\mathbb{E}_{z, y', x} [\Vert G_{y'}(z;\beta)-E_{y'}(T(x, G_{y'}(z;\beta);\alpha);\phi) \Vert_1].
\end{split}
\end{align}
To further stabilize the cooperative training and accelerate the MCMC teaching effect, we propose to add the following energy-based regularization on the translator, 
\begin{equation}
\label{eq:energy_generator_loss} \mathcal{L}_{\text{mode}}(\alpha,\beta) = \mathbb{E}_{z, y', x} [ D_{y'}(T(x, G_{y'}(z;\beta);\alpha);\theta) ],
\end{equation}
which can shift the translator mapping toward the low energy modes of the energy function.

\subsection{Cooperative Learning of Descriptor and Translator}

Our full objective function of the descriptor is $\mathcal{L}_{\text{descriptor}}=\mathcal{L}_\text{ebm}+\lambda_\text{energy} \mathcal{L}_\text{energy}$ and the full objective function of the translator is $\mathcal{L}_{\text{translator}}=\mathcal{L}_\text{teach}+\lambda_\text{diverse} \mathcal{L}_\text{diverse}+ \lambda_\text{cycle}\mathcal{L}_\text{cycle}+\lambda_\text{style}\mathcal{L}_\text{style}+\lambda_\text{mode}\mathcal{L}_\text{mode}$, where $\lambda_\text{energy}$, $\lambda_\text{diverse}$, $\lambda_\text{cycle}$, $\lambda_\text{style}$, and $\lambda_\text{mode}$ are hyperparameters. At each learning iteration, the cooperative learning algorithm alternates the following steps: (1) Generate an initial translated image via $\hat{x}=T(x,G_{y}(z))$; (2) Revise $\hat{x}$ by Langevin dynamics in Eq.\ref{eq: langevin_sampling} to obtain $\tilde{x}$;  (3) Update the parameters $\theta$ of descriptor by minimizing $\mathcal{L}_{\text{descriptor}}$; (4) Update the parameters $\alpha, \phi, \beta$ of translator by minimizing $\mathcal{L}_{\text{translator}}$.

\begin{algorithm}[H]
\begin{flushleft}
    \textbf{Input:} 
    Multi-resolution data  $\{(x^{(s)}_{i},y^{(s)}_{i})$, $i=1,...,N;s=1, ...,S\}$ 
    \textbf{Output:} Model $E^{(S)}, T^{(S)},D^{(S)},G$
\end{flushleft}
\begin{algorithmic}
\State $E^{(0)}\gets \varnothing,D^{(0)}\gets \varnothing,T^{(0)} \gets \varnothing$
\For {$s = 1, \cdots, S$}
\State $m \gets 0$
\IfThenElse {$s=1$}{$\omega \gets 1$}{$\omega \gets 0$}%
\State $E^{(s,\omega)} \gets \text{expand}(E^{(s-1)})$
\State $D^{(s,\omega)} \gets \text{expand}(D^{(s-1)})$
\State $T^{(s,\omega)} \gets \text{expand}(T^{(s-1)})$
\While {($m \leq N$)}
    \State Sample $(x,y)$ and  $y'$
    \State Sample $z \sim \mathcal{N}(0,I)$
    \State $c \gets E_{y}^{(s,\omega)}(x)$ or $c \gets G_{y'}(z)$
    \State $\hat{x} \gets T^{(s,\omega)}(x, c)$
    \State Revise $\hat{x}$ to obtain $\tilde{x}$ by a $K$-step Langevin dynamics in Eq.~(\ref{eq: langevin_sampling}).
    \State Update descriptor $D^{(s,\omega)}$ with $\mathcal{L}_{\text{descriptor}}$ 
    \State Update translator $\{E^{(s,\omega)},T^{(s,\omega)}, G\}$ with $\mathcal{L}_{\text{translator}}$
    \State $m \gets m+n^{(s)}$
    \IfThenElse {$s \neq 1$}{$\omega \gets \min(1,m/N)$}{$1$}%

\EndWhile
\EndFor

\end{algorithmic}
\caption{Progressive Cooperative Learning}
\label{alg:scale}
\end{algorithm}

\begin{figure*}[ht]
    \centering
    \includegraphics[width=\textwidth]{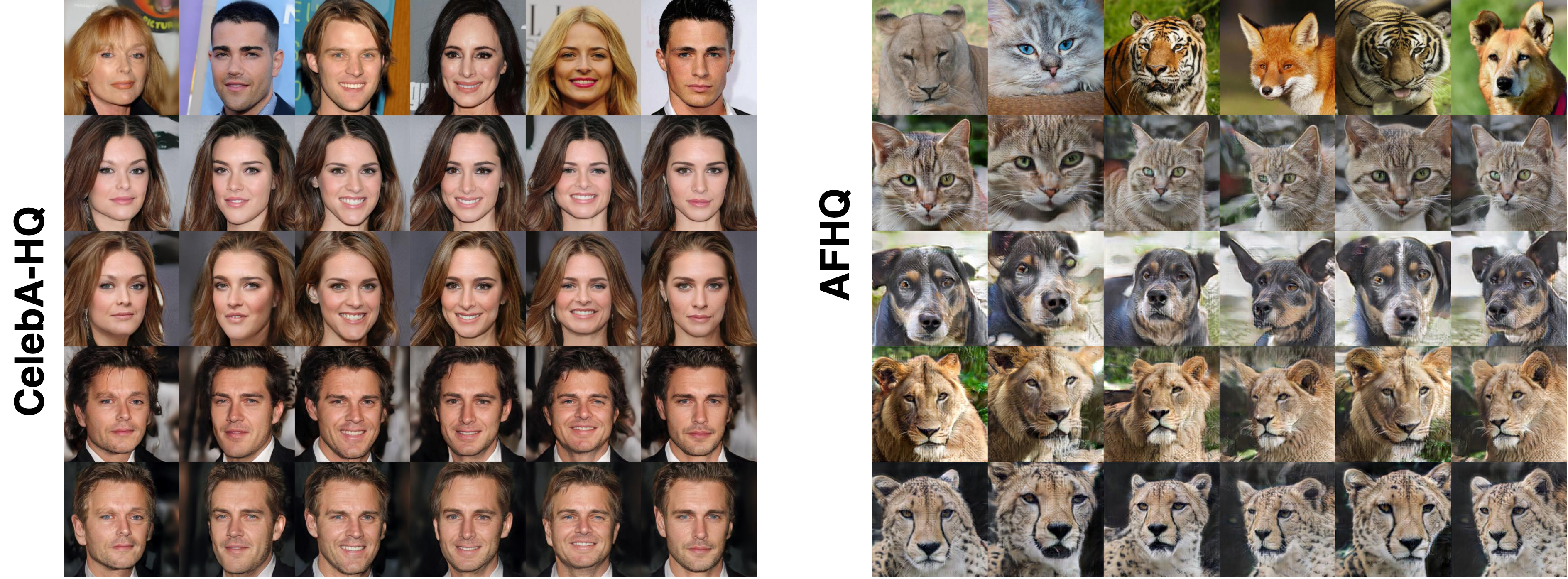}
    \caption{Qualitative results of diverse image generation for human face on CelebA-HQ dataset (left) and animal face on AFHQ dataset (right) are shown in this figure. Each column displays one example of one-to-many image generation. The first row displays source images. The rest four rows show different translated images, which are obtained by using four style codes randomly generated by the style generator. The style generator produces style codes by randomly sampling from Gaussian distribution.}
    \label{fig:exp_diverse}
\end{figure*}

\subsection{Progressive Cooperative Learning}
The update of both descriptor and translator relies on the cooperative generation of MCMC synthesized examples, denoted as $\tilde{x}$. To significantly improve training efficiency, we propose a progressive learning strategy for our cooperative learning framework. The algorithm gradually enhances the model resolution from low to high, while maintaining cooperative learning across all components at each resolution. The underlying motivation behind this strategy is that learning and sampling from a low resolution data domain is much more efficient. By leveraging a pre-trained low resolution model as a foundation, we can efficiently learn the next scale of the model, rather than starting from scratch. When expanding the current model to the next scale, each component's network structure undergoes modifications. New layers are added to handle higher resolution image inputs or outputs, while incompatible old layers are removed. The newly added layers are trained together with the remaining parameters. To ensure a smooth transition and prevent gradient exposure due to the addition of expanding blocks in each component, we propose to retain partial effects of the parameters that need to be removed while incorporating the effects of the newly added parameters. Throughout each resolution of learning, the impact of the removed parameters gradually diminishes until it becomes zero. Figure~\ref{fig:psc_progressive} illustrates the expanding strategy of each component at every level of resolution. Here, $\omega$ represents a transition factor that starts from 0 and increase to 1, controlling the percentage of effects from the parameters to be removed (depicted as dark grey boxes with dashed boundaries) and the parameters to be added (depicted as light grey boxes). For a complete description of the proposed progressive cooperative learning algorithm, please refer to Algorithm \ref{alg:scale}.
\section{Experiment}

\subsection{Experiment Settings}

\paragraph{Dataset and Evaluation Metrics}
To demonstrate the performance of our proposed multi-domain image-to-image translation framework, we test it on the CelebA-HQ \cite{progressive_gan} and AFHQ \cite{stargan2} datasets and compare it with several baselines. We use M and F to refer the domains of male and female in CelebA-HQ, and C, D and W to refer to the domains of cat, dog, and wild animals in AFHQ. We only use the images and the corresponding domain labels from the datasets in our experiments. We evaluate the quality of translated images using the Fréchet Inception Distance (FID) \cite{fid} and the Kernel Inception Distance (KID) \cite{binkowski2018demystifying}, which are widely used to measure the distance between the population of translated images and the population of original images in the target domain. A small FID or KID is desired to indicate that the translated distribution is very close to the target distribution.

\begin{figure*}[ht]
    \centering
    \includegraphics[width=0.95\textwidth]{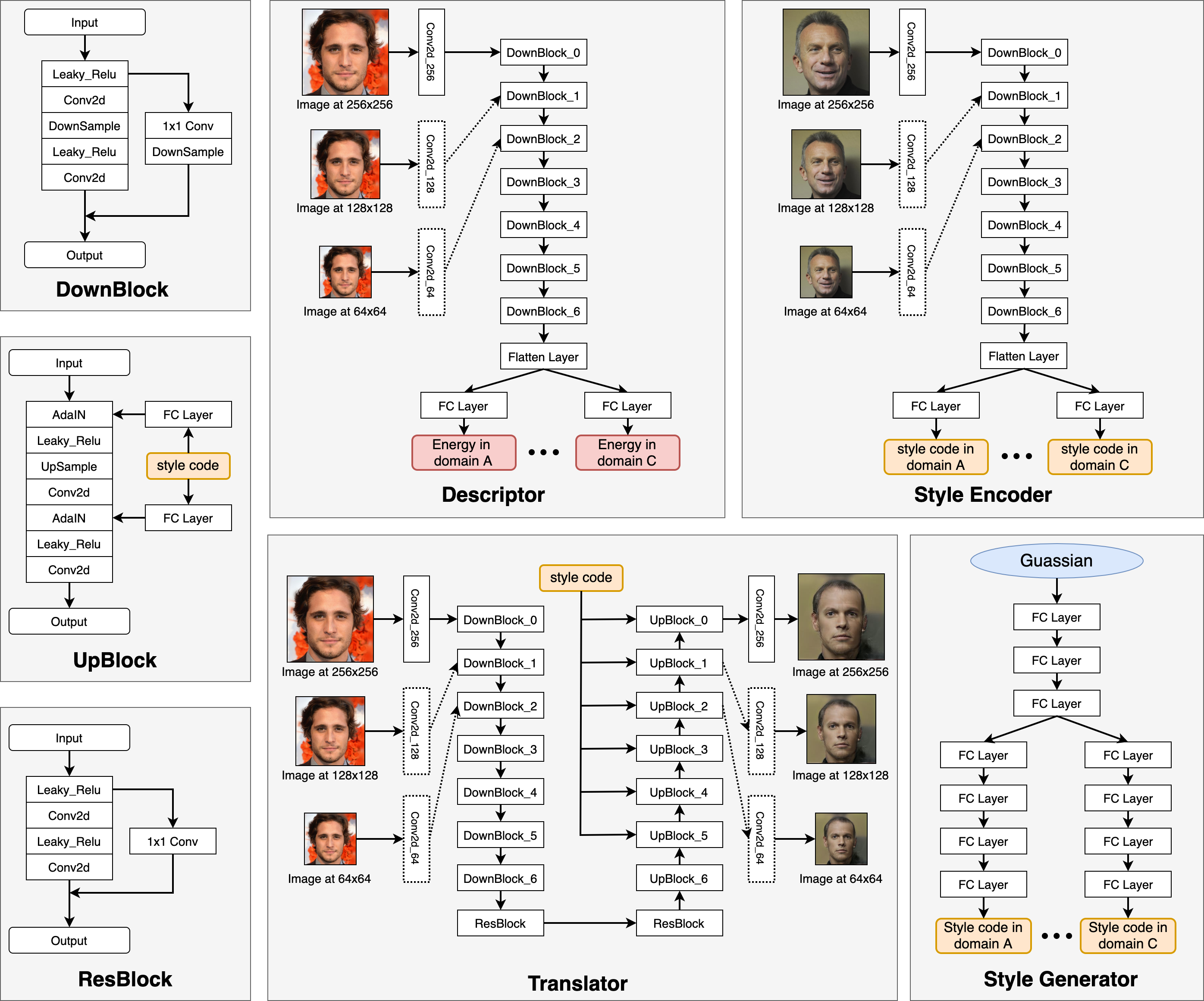}
    \caption{Architecture of proposed networks}
    \label{fig:architecture}
\end{figure*}

\paragraph{Network Architecture}
We show the detail of network structures in Fig.~\ref{fig:architecture}. We use dotted lines to represent the layers that will be dropped off during scaling up in Descriptor, Translator, and Style Encoder. Since the Style Generator samples from Guassian distrbution, the architecture keeps the same during progression. For each DownBlock/UpBlock module at level $l$, we set the input and out channel to be $max(512, 2^{5+l})$ and $max(512, 2^{6+l})$. For the FC Layer in Style Generator, we set the input channel of the top FC Layer as 16 and the rest as 512. The size of the style code is set to be 64 for both experiments.

\begin{figure*}[ht]
    \centering
    \includegraphics[width=0.9\textwidth]{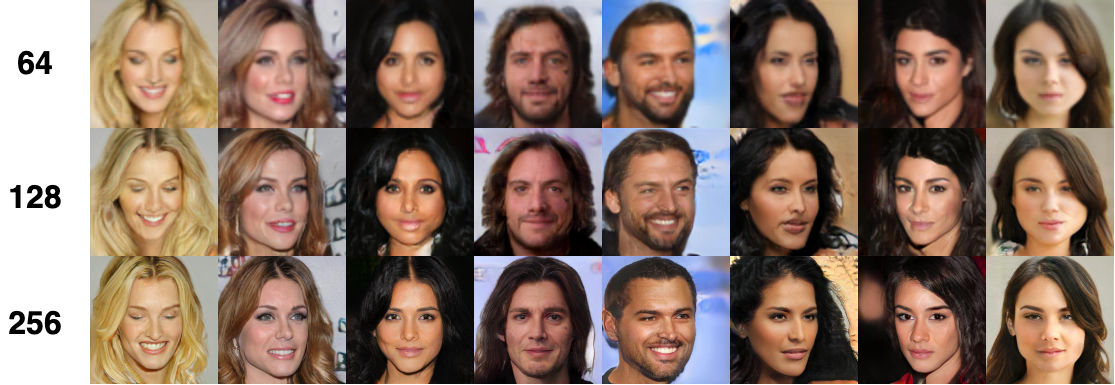}
    \caption{Generation results on human face in different resolution.}
    \label{fig:celeba_progressive_supp}
\end{figure*}

\begin{figure*}[ht]
    \centering
    \includegraphics[width=0.9\textwidth]{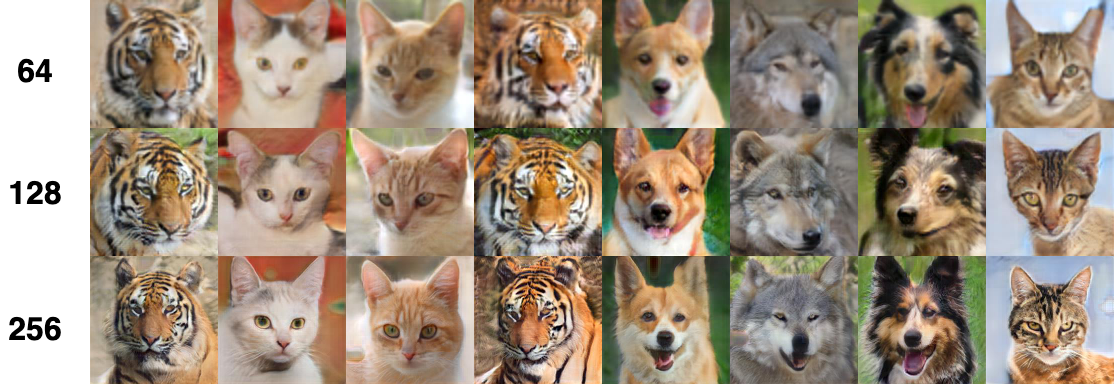}
    \caption{Generation results on animal face in different resolution.}
    \label{fig:afhq_progressive_supp}
\end{figure*}

\paragraph{Progressive Training}
We start training our model with a resolution of 64$\times$64, and then scale it up to 128$\times$128 and 256$\times$256. Example results after each progression for human face generation in Figure \ref{fig:celeba_progressive_supp} and animal face generation in Figure \ref{fig:afhq_progressive_supp}. Consistently, we could see the generation quality improves after each step of progression. We step for 16 iterations for MCMC sampling in the beginning, decreasing by 4 steps after each progression. The hyper-parameters of $\lambda_\text{energy}$, $\lambda_\text{diverse}$, $\lambda_\text{cycle}$, $\lambda_\text{style}$, and $\lambda_\text{mode}$ are set to be 1.

\begin{figure*}[ht]
    \centering
    \includegraphics[width=0.9\textwidth]{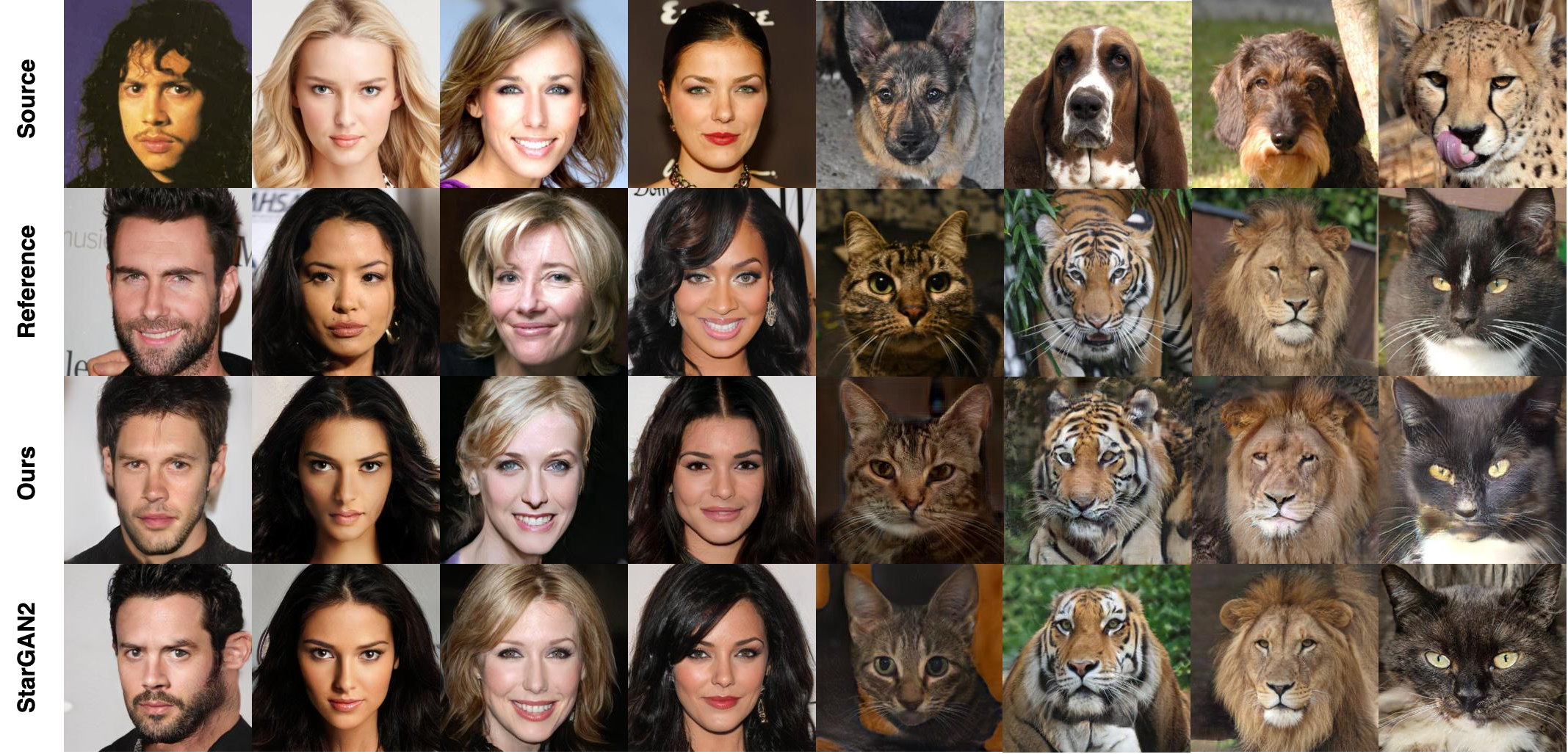}
    \caption{Comparison of human face and animal face translation results between EBM and GAN.}
    \label{fig:compare}
\end{figure*}

\begin{table*}[t]
    \centering
    \caption{Evaluation on CelebA-HQ dataset for two-domain human face generation and AFHQ dataset dataset for three-domain animal face generation.}
    \label{tab:compare_multidomain}
    \renewcommand{\arraystretch}{1.2}
    \begin{tabular}{p{3.0cm}<{\centering}p{1.1cm}<{\centering}p{1.1cm}<{\centering}p{0.1cm}p{1.1cm}<{\centering}p{1.1cm}<{\centering}p{0.1cm}p{1.1cm}<{\centering}p{1.1cm}<{\centering}p{0.1cm}<{\centering}p{1.1cm}<{\centering}p{1.1cm}<{\centering}}
    \hline
    \multirow{3}{*}{Method}&\multicolumn{5}{c}{Reference}&&\multicolumn{5}{c}{Diverse}\cr
    \cline{2-6}\cline{8-12}
    &\multicolumn{2}{c}{CelebA-HQ}&&\multicolumn{2}{c}{AFHQ}&&\multicolumn{2}{c}{CelebA-HQ}&&\multicolumn{2}{c}{AFHQ}\cr
    \cline{2-3}\cline{5-6}\cline{8-9}\cline{11-12}
    &FID&KID&&FID&KID&&FID&KID&&FID&KID\cr
    \hline
    MUNIT\cite{munit}&107.1&-&&223.9&-&&31.4&-&&41.5&-\cr
    DRIT\cite{drit}&53.3&-&&114.8&-&&52.1&-&&95.6&-\cr
    MSGAN\cite{ms_gan}&39.6&-&&69.8&-&&33.1&-&&61.4&-\cr
    StarGAN2\cite{stargan2}&23.8&12.1&&19.8&6.1&&13.7&4.1&&16.2&9.1\cr
    Liu\cite{liu2021smoothing}&26.7&16.8&&51.7&28.6&&17.8&11.0&&26.0&7.0\cr
    TUNIT\cite{baek2021rethinking}&173.7&193.7&&223.0&187.7&&128.0&122.0&&116.1&99.7\cr
    SwapAE\cite{park2020swapping}&25.4&17.8&&61.2&28.8&&-&-&&-&-\cr
    CLUIT\cite{lee2021contrastive}&28.9&18.1&&22.6&10.5&&-&-&&-&-\cr
    StyleMapGAN\cite{kim2021exploiting}&28.8&25.1&&64.3&51.3&&24.3&15.2&&32.8&18.7\cr
    CycleCoop\cite{coop_cycle}&-&-&&-&-&&131.0&124.7&&-&-\cr
    EM-LAST\cite{han2022last}&-&-&&-&-&&48.8&22.9&&41.5&17.0\cr
    \textbf{Ours}&\textbf{21.0}&\textbf{7.7}&&21.0&7.9&&32.9&21.9&&31.8&16.9\cr
    \hline
    \end{tabular}
\end{table*}

\begin{table*}[tp]
    \centering
    \caption{Ablation Study on CelebA-HQ and AFHQ datasets in 64$\times$64 resolution.}
    \label{tab:exp_abalation}
    \renewcommand{\arraystretch}{1.2}
    \begin{tabular}{p{3.2cm}<{\centering}p{1.4cm}<{\centering}p{1.4cm}<{\centering}p{0.1cm}p{1.4cm}<{\centering}p{1.4cm}<{\centering}p{1.4cm}<{\centering}}
    \hline
    \multirow{2}{*}{Removed Item}&\multicolumn{2}{c}{CelebA}&&\multicolumn{2}{c}{AFHQ}&\multirow{2}{*}{Avg}\cr
    \cline{2-3}\cline{5-6}
    &Reference&Diverse&&Reference&Diverse\cr
    \hline
    baseline&15.1&14.3&&12.4&19.6&15.4\cr
    Remove {$\mathcal{L}_{diverse}$}&16.3&17.1&&36.4&35.2&26.3\cr
    Remove {$\mathcal{L}_{cycle}$}&111.0&127.3&&NA&NA&119.2\cr
    Remove {$\mathcal{L}_{energy}$}&134.5&40.7&&208.5&97.6&120.3\cr
    Remove {$\mathcal{L}_{mode}$}&NA&NA&&277.2&217.6&247.4\cr
    \hline
    \end{tabular}
\end{table*}


\begin{table}[ht]
    \centering
    \caption{Evaluation on specific domain translations by FID score.}
    \label{tab:compare_specificdomain}
    \renewcommand{\arraystretch}{1.2}
    \begin{tabular}{p{2.8cm}<{\centering}p{1.2cm}<{\centering}p{1.2cm}<{\centering}p{1.2cm}<{\centering}p{1.2cm}<{\centering}}
    \hline
    Method&C$\rightarrow$D&W$\rightarrow$D&M$\rightarrow$F\cr
    \hline
    ILVR\cite{choi2021ilvr} &74.4&75.3&46.1\cr
    SDEdit\cite{meng2021sdedit}&74.2&68.5&49.4\cr
    CUT\cite{park2020contrastive}&76.2&92.9&31.9\cr
    C2F-EBM\cite{ebm_c2f}&55.1&-&-\cr
    EM-LAST\cite{han2022last}&69.4&72.5&47.8\cr
    EGSDE\cite{zhao2022egsde}&51.0&50.4&30.6\cr
    \textbf{Ours} (Diverse)&55.1&54.5&26.8\cr
    \textbf{Ours}(Reference)&\textbf{42.2}&\textbf{41.8}&\textbf{16.1}\cr
    \hline
    \end{tabular}
\end{table}

\begin{table}[tp]
    \centering
    \caption{Complexity analysis of different models. The units for training and inference (including diverse and reference generation) are s/batch and ms/batch.}
    \begin{tabular}{p{1.6cm}<{\centering}p{1.6cm}<{\centering}p{1.6cm}<{\centering}p{1.6cm}<{\centering}}
    \hline
    Method&Ours&StarGAN2&CycleCoop\cr
    \hline
    Param(M)&87.7&87.7&108.4\cr
    Training&2.7&1.0&2.7\cr
    Reference&87.08&86.86&NA\cr
    Diverse&20.17&21.22&9.55\cr
    \hline
    \end{tabular}
    \label{tab:complexity}
\end{table}

\begin{figure*}[ht]
    \centering
    \includegraphics[width=\textwidth]{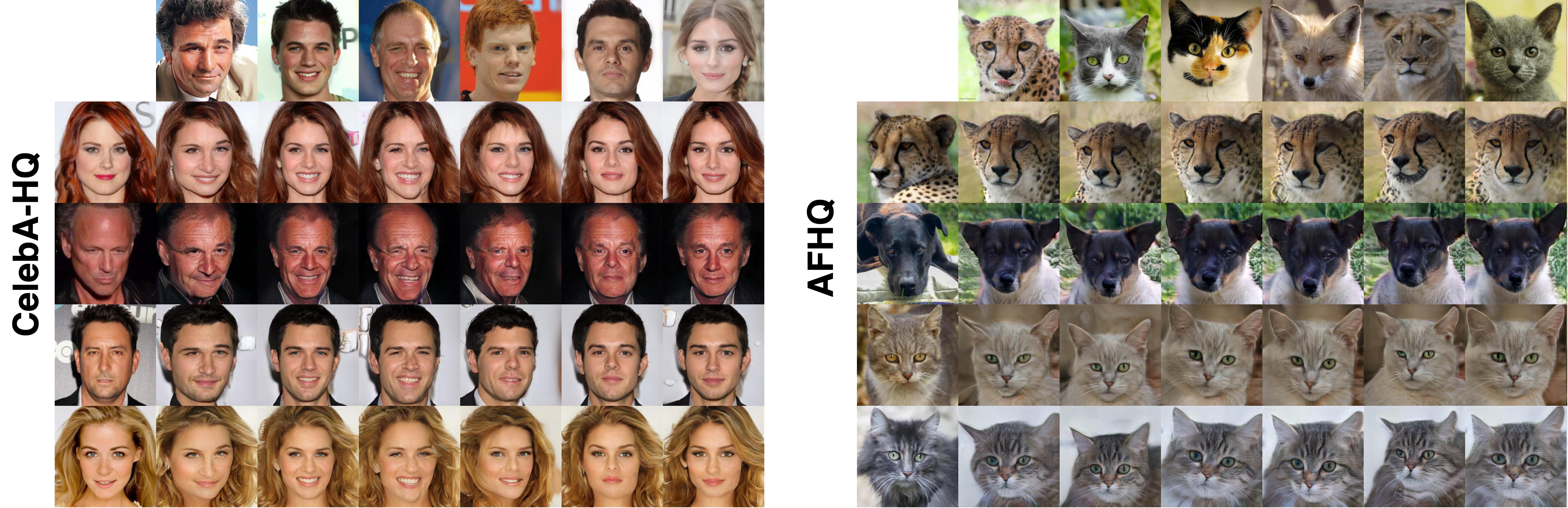}
    \caption{We show the translated images with style codes generated from the Style Encoder and reference images for human (left) and animal (right) face in this picture. The source images and reference images are put in the first row and first column. We could see that the face has successfully translated into target domains with consistency in expression.}
    \label{fig:exp_reference}
\end{figure*}

\subsection{Diverse Image Generation}
 In this experiment, we use style codes that are randomly sampled from the style generator to generate diverse translated images. Examples of generation results for human face on the CelebA-HQ dataset and animal face on the AFHQ dataset can be seen in Figure~\ref{fig:exp_diverse}. For each source image shown in the first row, we generate multiple outputs using random Gaussian noise. The qualitative results verify the diversity of the translated results from a source input image. We observe that, given a source image, our model can not only generate diverse translated images but also produce high-quality images that obtain the same attribute (e.g., expression) from the source image.

\subsection{Translation with Reference Image}
We perform image-to-image translation by providing a reference image. We first adopt the style encoder to extract the style code from the provided reference image, and the apply the style code to the translator. 
Figure~\ref{fig:exp_reference} show some qualitative results, where we take images in the first row as source images and images in the first column as reference images. The translation results are shown in the middle. Comparing results displayed in each row, we can observe that the human face in the source image can be clearly changed into the same gender and appearance of the face in the reference image, while keeping the facial expression consistent with that in the source domain.




\subsection{Quantitative Comparison}
We also compare the results of our translation results quantitatively by using style codes from Style Encoder by randomly selecting reference image in different domains or from Style Generator through sampling from Gaussian distribution with other baseline methods based on adversarial learning, score matching, or EBMs quantitatively. For each source image in the validation dataset, we obtain ten translated images for each target domain to compute the FID. Results for both human and animal face translation are shown in Table\ref{tab:compare_multidomain}. We also compare our results with some pair-wise translation models on specific domain transfer and summarize the results in Table \ref{tab:compare_specificdomain}. We could see that our model could significantly out perform existing cooperative learning methods with additional ability of guidance by reference images and reach comparable performance with GAN-based methods.

\subsection{Comparison with Adversarial Learning}
We have a comparison of generation results from EBM (ours) and GAN (StarGAN2) in human and animal face translation with reference images and show the results in Figure \ref{fig:compare}. Consistently with quantitative results of Table \ref{tab:compare_multidomain}, our model could generate comparable translation results in high resolution against GAN-based models.

\subsection{Complexity Analysis}
We have a comparison in computational cost below with the evaluation on a single Nvidia A100 GPU and show the results in Table \ref{tab:complexity}. We set the training batch to be 8 and inference batch to be 32 with resolution at 256 $\times$ 256 for all methods. CycleCoop is an EBM baseline only for two domains. StarGANv2 is the GAN-based baseline.

\subsection{Ablation Study}
We conduct an ablation study to evaluate the importance of each individual component proposed in our paper. In Table \ref{tab:exp_abalation}, we report the model performance in terms of FID by removing different key loss term (including $\mathcal{L}_{\text{diverse}}, \mathcal{L}_{\text{cycle}},\mathcal{L}_{\text{energy}},\mathcal{L}_{\text{mode}}$) from our objective function in our framework. We train our model in a 64$\times$64 resolution setting on datasets CelebA-HQ and AFHQ without using the progressive learning strategy. We show results of image translation using style codes obtained from both style encoder and style generator and report average performance. NA means that the model fails in learning and can not generate meaningful results. We can see that the newly added regularization strategies for the descriptor and the translator, i.e., $\mathcal{L}_{energy}$ and $\mathcal{L}_{mode}$, are essential for stabilizing the cooperative training. Especially, the energy-based regularization loss $\mathcal{L}_{mode}$ plays an important role to ensure that the translator can quickly catch up with the descriptor toward the data distribution during the cooperative training. The $\mathcal{L}_{energy}$ is useful to obtain performance gain by limiting the magnitude of the energy values. Also, we can find that the performance drops significantly when removing the cycle-consistency loss $\mathcal{L}_{cycle}$, which proves to be a key objective for unpaired cross-domain image translation task.

\section{Conclusion}
We present PMD-CoopNets, a novel approach that combines energy-based learning, MCMC sampling, cooperative learning, and progressive learning for unpaired multi-domain image-to-image translation. Our method includes a multi-head energy-based model as a descriptor, capturing the multi-domain image distribution, and a diversified image-to-image translator for cross-domain one-to-many mapping. To train both the descriptor and translator, we introduce a multi-domain MCMC teaching algorithm. Additionally, we propose progressive learning to enhance the scalability and efficiency of our model. Experimental results demonstrate that our approach achieves comparable performance to adversarial learning frameworks and sets a new benchmark in energy-based image-to-image translation methods. 

\clearpage
{
    \small
    \bibliographystyle{ieeenat_fullname}
    \bibliography{main}
}

\end{document}


\maketitle

\section{Network Architectures}
\begin{figure*}[ht]
    \centering
    \includegraphics[width=\textwidth]{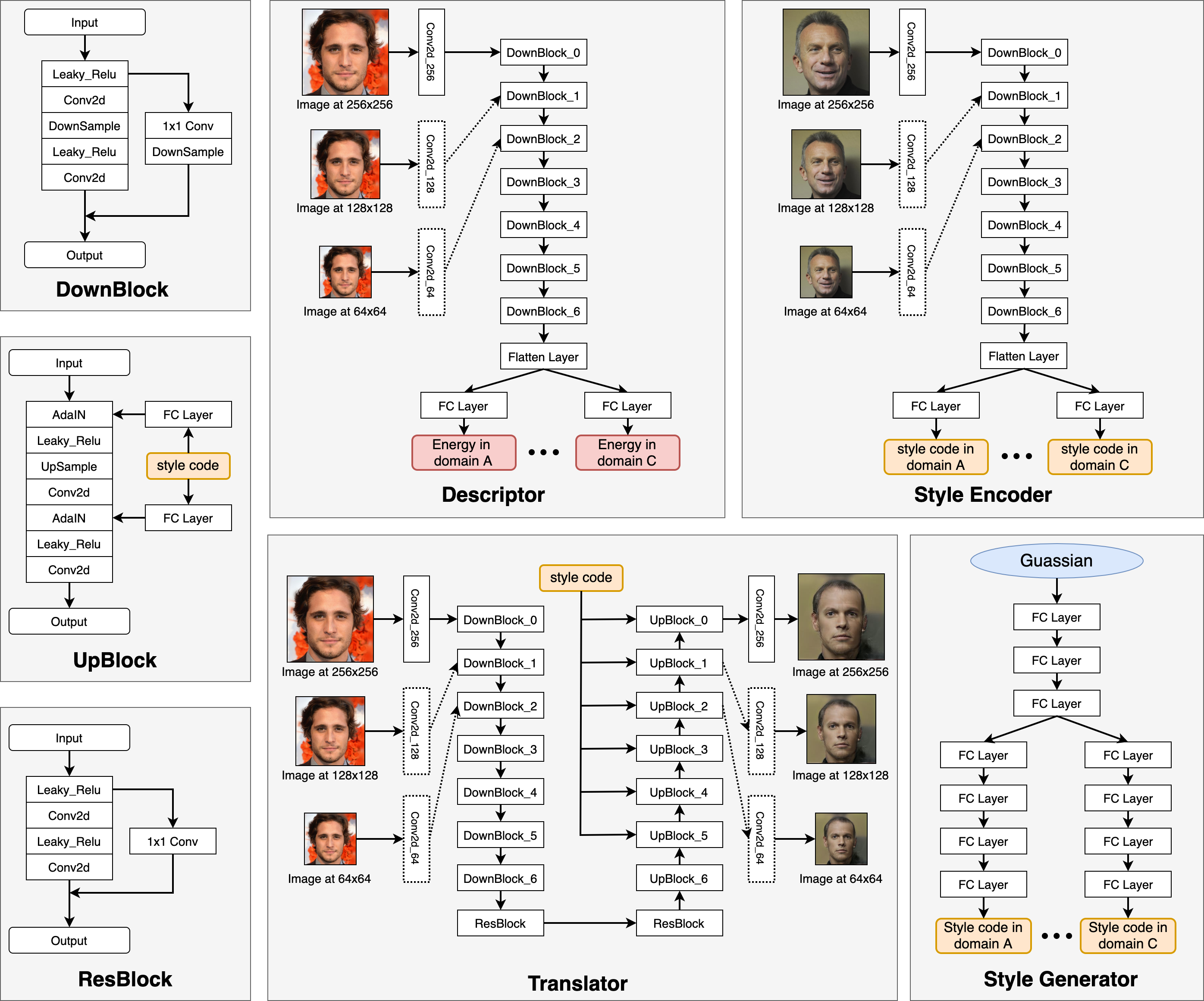}
    \caption{Architecture of proposed networks}
    \label{fig:architecture}
\end{figure*}

We show the detail of network structures in Fig.~\ref{fig:architecture}. We use dotted lines to represent the layers that will be dropped off during scaling up in Descriptor, Translator, and Style Encoder. Since the Style Generator samples from Guassian distrbution, the architecture keeps the same during progression. For each DownBlock/UpBlock module at level $l$, we set the input and out channel to be $max(512, 2^{5+l})$ and $max(512, 2^{6+l})$. For the FC Layer in Style Generator, we set the input channel of the top FC Layer as 16 and the rest as 512. The size of the style code is set to be 64 for both experiments.

\section{Cycle Translation}
To check the cycle consistency of our model, we first perform an image-to-image translation from source domain to target domain, and then perform a reverse translation by treating the translated image as the source image and translating it back to the original domain. Qualitative results for human and animal face translation are displayed in Fig~\ref{fig:celeba_cyc} and Fig~\ref{fig:afhq_cyc}. The qualitative results show that our model could consistently keep content structure during translation.

\begin{figure*}[ht]
    \centering
    \includegraphics[width=\textwidth]{supplemental_figs/celeba_cyc.png}
    \caption{Cycle translation on human face.}
    \label{fig:celeba_cyc}
\end{figure*}

\begin{figure*}[ht]
    \centering
    \includegraphics[width=\textwidth]{supplemental_figs/afhq_cyc.png}
    \caption{Cycle translation on animal face.}
    \label{fig:afhq_cyc}
\end{figure*}

\section{Progressive Generation Results}
\begin{figure*}[ht]
    \centering
    \includegraphics[width=\textwidth]{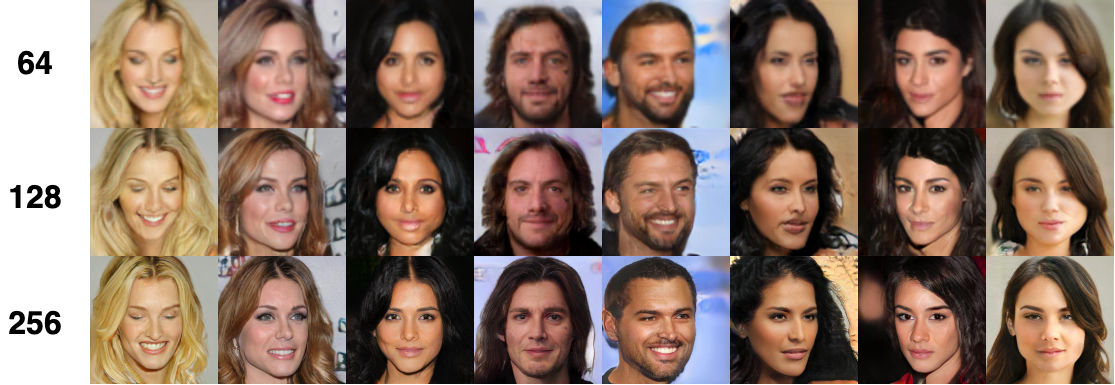}
    \caption{Generation results on human face in different resolution.}
    \label{fig:celeba_progressive_supp}
\end{figure*}

\begin{figure*}[ht]
    \centering
    \includegraphics[width=\textwidth]{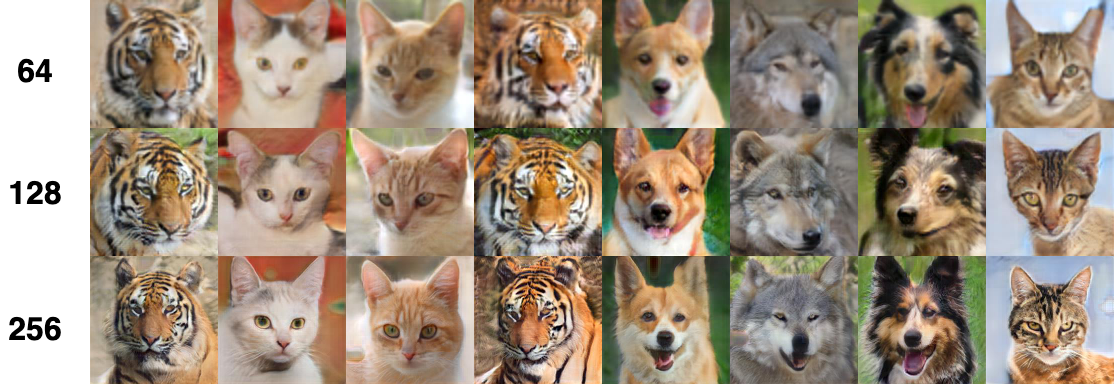}
    \caption{Generation results on animal face in different resolution.}
    \label{fig:afhq_progressive_supp}
\end{figure*}

We break the training process into stage, starting from the resolution at $64\times64$ and scale up the Translater, StyleEnocder, and Descriptor to $128\times128$ and $256\times256$ consecutively. We show additional results at the end of each progression step for human face generation in Figure \ref{fig:celeba_progressive_supp} and animal face generation in Figure \ref{fig:afhq_progressive_supp}. Consistently, we could see the generation quality improves after each step of progression.

\section{Comparison with Adversarial Learning}
We have a comparison of generation results from EBM (ours) and GAN (StarGAN2) in human and animal face translation with reference images and show the results in Figure \ref{fig:celeba_compare} and Figure \ref{fig:afhq_compare}. Consistently with quantitative results of Table 1 in the main paper, our model could generate comparable translation results in high resolution against GAN-based models.

\begin{figure*}[ht]
    \centering
    \includegraphics[width=\textwidth]{supplemental_figs/celeba_compare.png}
    \caption{Comparison of human face translation results between EBM and GAN.}
    \label{fig:celeba_compare}
\end{figure*}

\begin{figure*}[ht]
    \centering
    \includegraphics[width=\textwidth]{supplemental_figs/afhq_compare.png}
    \caption{Comparison of animal face translation results between EBM and GAN.}
    \label{fig:afhq_compare}
\end{figure*}

\section{Complexity Analysis}
We have a comparison in computational cost below with the evaluation on a single Nvidia A100 GPU and show the results in Table \ref{tab:complexity}. We set the training batch to be 8 and inference batch to be 32 with resolution at 256 $\times$ 256 for all methods. CycleCoop is an EBM baseline only for two domains. StarGANv2 is the GAN-based baseline.

\begin{table*}[ht]
    \centering
    \begin{tabular}{ccccc}
    \hline
    Method&Param(M)&Training (s/batch)&Reference Inference (ms/batch)&Diverse Inference (ms/batch)\cr
    Ours&87.7&2.7&87.08&20.17\cr
    StarGAN2&87.7&1.0&86.86&21.22\cr
    CycleCoop&108.4&2.7&NA&9.55\cr
    \hline
    \end{tabular}
    \caption{Complexity Analysis}
    \label{tab:complexity}
\end{table*}
